# A TWO-PASS FUZZY-GENO APPROACH TO PATTERN CLASSIFICATION

*Subhadip Basu (subhadip8@yahoo.com), Mahantapas Kundu (mkundu2000@yahoo.co.in) *,*
*Mita Nasipuri (nasipuri@vsnl.com) * and Dipak Kumar Basu (dipakbasu@hotmail.com) **

*Computer Science & Engineering Department, MCKV Institute of Engineering,
Liluah, Howrah-711204, India.*
**Computer Science & Engineering Department, Jadavpur University,
Kolkata-700032, India.*

*Abstract:*
*The work presents an extension of the fuzzy approach to 2-D shape recognition [1] through refinement of initial or coarse classification decisions under a two pass approach. In this approach, an unknown pattern is classified by refining possible classification decisions obtained through coarse classification of the same. To build a fuzzy model of a pattern class horizontal and vertical fuzzy partitions on the sample images of the class are optimized using genetic algorithm. To make coarse classification decisions about an unknown pattern, the fuzzy representation of the pattern is compared with models of all pattern classes through a specially designed similarity measure. Coarse classification decisions are refined in the second pass to obtain the final classification decision of the unknown pattern. To do so, optimized horizontal and vertical fuzzy partitions are again created on certain regions of the image frame, specific to each group of similar type of pattern classes. It is observed through experiments that the technique improves the overall recognition rate from 86.2%, in the first pass, to 90.4% after the second pass, with 500 training samples of handwritten digits.*

## 1.0 INTRODUCTION

The two-pass classifier, described under the present work, first performs a *coarse classification* on the input pattern by restricting the possibility of classification decision within a group of classes, smaller than the original group of classes considered initially. In the second pass, the classifier refines its earlier decision by selecting the true class of the input pattern from the group of candidate classes selected in the first pass. In doing so, unlike the previous pass, the classifier concentrates only on certain regions of the input pattern, specific to the group of classes selected in the earlier pass. The group of candidate classes formed in the first pass of classification is determined by the *top choice* of the classifier in the same pass. There is a high chance that an input pattern classified into a member class of such group originally belongs to some other member class of the same group. By observing the *confusion matrix* on the *training data*, all such necessary groups of candidate classes can be formed for a particular application. The groups are formed on the basis of the statistical information obtained through the application of the classifier on the training data with the same features selected for the first pass. Secondary choices of the classifier are not considered in selection or for formation of a group.

The technique presented here is applied for Optical Character Recognition (OCR) related applications. OCR systems appear to ease the barrier of the keyboard interface between man and machine, and help in office automation with huge saving of time and human effort. The OCR systems have potential applications in extracting data from filled in forms, interpreting handwritten addresses from postal documents for automatic routing, automatic reading of bank cheques etc.

The work presented here embodies results of an investigation of the proposed technique by experimenting with the *handwritten digit recognition* problem. Handwritten digit recognition is a realistic benchmark problem of pattern recognition. In the present work we have developed a novel technique for recognition of handwritten Bangla digits using a two-pass approach. Fig. 1 shows the typical *Bangla* digit patterns of all ten decimal digits in ascending order of values.

**Fig. 1.** The decimal digit set of Bangla script

The recent trend for improving the performance of a pattern recognition system is to combine the complementary information provided by multiple classifiers [2-5]. The digit patterns considered here consist of samples of handwritten digits of *Bangla* script. Popularity wise, *Bangla* stands 2$^{nd}$ after Hindi, both as a script and a language in Indian subcontinent. Compared to OCR researches on Roman scripts [1-4, 6-8], *Bangla* has received little attention as a subject of OCR research [5, 9] until recently.

**2.0 FEATURE SELECTION**

To design the feature set during coarse classification, we have introduced horizontal and vertical fuzzy partitions on each pattern image. These partitions are optimized using genetic algorithm to give best possible recognition performances. In this work, images of handwritten digit samples are each scaled to 32x32 pixels size to ease the feature extraction process. Each of these images is divided into a number of sub-images through the fuzzy partitions for extraction of fuzzy features. The size and coordinates of these sub-images are encoded in the genes of a single chromosome. Each chromosome has two parts. The first part consists of horizontal partition coordinates of each sub-image and the second part of the chromosome contains the coordinates for the vertical partitions. A sample chromosome structure, that encodes the coordinate values in its gene, is shown below. Fig. 2 shows the corresponding horizontal and vertical partitions of the image frame.

           **Part 1**         **Part 2**
Ch [0] = 0, 7, 9, 13, 21, 23, 31 | 0, 11, 19, 24, 31

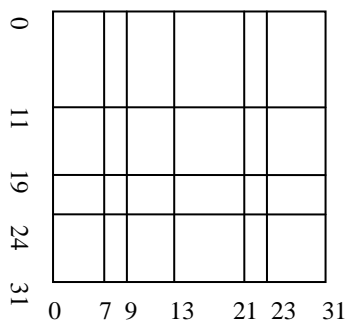

**Fig. 2.** Formation of sub-images from a chromosome

Here, each chromosome is an integer valued, two part, variable length chromosome. The initial population is a collection of random valued chromosomes. For a given chromosome, all the training patterns are divided in sub-images according to the gene values of the same. The fitness function is designed to compute an overall success rate for all the training patterns for each chromosome. Details of the fitness function, mutation-crossover strategy, termination condition for the genetic algorithm and the membership function for calculation of features in each sub-image window are discussed in [1]. Over the generations, these chromosomes evolve to produce better and better partitions for the digit images. When the termination condition is achieved, the top choice of the population gives the optimum partitioning of the training patterns to give the best possible recognition performance in coarse classification.

### 3.0 FORMATION OF GROUPS FOR FINER CLASSIFICATION

In the present work, we have extended the concept, developed by Lazzerini et.al. in [1], for finer classification of digit patterns of handwritten Bangla script. To generate optimum fuzzy partitions during coarse classification, 50 samples each are taken from the 10 digit samples of handwritten Bangla script. The recognition rate after coarse classification is 86.2%. The confusion matrix, so obtained after optimum partitioning of the 500 digit samples, is shown in Fig. 3.

```
       0  1  2  3  4  5  6  7  8  9
       --------------------------------
0 :   49  0  0  0  0  1  0  0  0  0
1 :    0 46  0  0  1  0  0  1  2  0
2 :    0  2 46  0  2  0  0  0  0  0
3 :    1  0  0 37  0  3  6  2  1  0
4 :    0  0  0  0 48  0  0  0  1  1
5 :    9  0  1  0  0 39  0  0  1  0
6 :    0  0  0 14  0  2 33  0  0  1
7 :    0  0  0  0  1  0  0 49  0  0
8 :    0  0  0  0  0  0  0  0 50  0
9 :    1  7  0  1  3  0  0  3  1 34
```

**Fig. 3.** Confusion matrix after coarse classification

From the matrix it is evident that there major misclassifications with pattern classes (3, 6), (0, 5) and (1, 9). To refine this result, finer classification is done on these three groups of digit patterns. Overlapped samples of ten different digit classes of *Bangla* script is shown in Fig. 4. This image is generated by overlapping all the normalized digit images of each class and stretching the pixel intensity between (0, 255). From these overlapped images it can be observed that there exists some confusion region within each group of patterns i.e. group (3, 6), (0, 5) and (1, 9). In each of these three groups, fuzzy partitioning is again performed to divide the confusion region into number of sub-images. An optimum partition is then obtained to produce best possible recognition performance within each of these three groups.

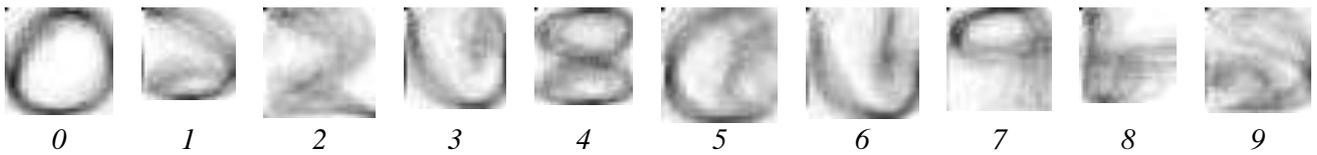

*0     1     2     3     4     5     6     7     8     9*

**Fig. 4.** Overlapped samples of ten different digit classes.

### 3.1. Finer Classification for Group (3, 6)

In group (3, 6) the top left and bottom right coordinates of the confusion region is identified as (0, 18) & (24, 31). Fig. 5 shows the overlapped images pattern classes of 3 and 6 with identified confusion regions. Based on this confusion region the classes 3 & 6 are classified. The recognition rate achieved within this group is 90%.

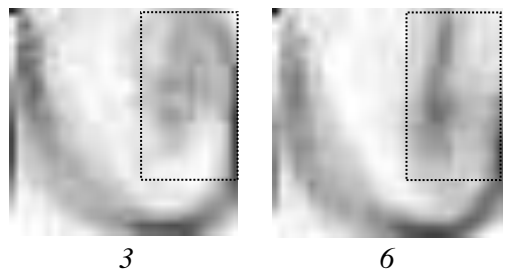

*3                    6*

**Fig. 5.** Overlapped digit patterns of group (3, 6) with identified confusion regions

### 3.2. Finer Classification for Group (0, 5)

In group (0, 5) the top left and bottom right coordinates of the confusion region is identified as (0, 15) & (31, 31). Fig. 6 shows the overlapped images pattern classes of 0 and 5 with identified confusion regions. Based on this confusion region the classes 0 & 5 are classified. The recognition rate achieved within this group is 96%.

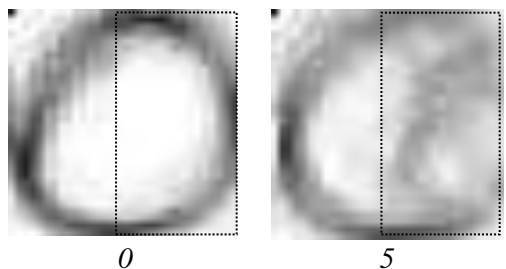

*0                    5*

**Fig. 6.** Overlapped digit patterns of group (0, 5) with identified confusion regions

### 3.3. Finer Classification for Group (1, 9)

In group (1, 9) the top left and bottom right coordinates of the confusion region is identified as (12, 0) & (31, 31). Fig. 7 shows the overlapped images pattern classes of 1 and 9 with identified confusion regions. Based on this confusion region the classes 1 & 9 are classified. The recognition rate achieved within this group is 95%.

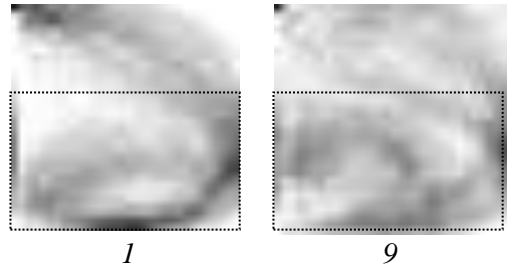

*1*　　　　*9*

**Fig. 7.** Overlapped digit patterns of group (1, 9) with identified confusion regions

## 4.0 RESULTS AND DISCUSSION

After coarse classification of an unknown pattern into one of the above groups, the final decision about its class membership is determined by considering features from the confusion regions of the said group. The overall success rate as observed by application of this two pass technique is 90.4%. Fig. 8 shows the confusion matrix obtained after second pass of classification on 500 training samples.

```
       0  1  2  3  4  5  6  7  8  9
       ----------------------------------
0 :50  0  0  0  0  0  0  0  0  0
1 : 0 44  0  0  1  0  0  1  2  2
2 : 0  2 46  0  2  0  0  0  0  0
3 : 1  0  0 40  0  3  3  2  1  0
4 : 0  0  0  0 48  0  0  0  1  1
5 : 4  0  1  0  0 44  0  0  1  0
6 : 0  0  0  5  0  2 42  0  0  1
7 : 0  0  0  0  1  0  0 49  0  0
8 : 0  0  0  0  0  0  0  0 50  0
9 : 1  2  0  1  3  0  0  3  1 39
```

**Fig. 8.** Confusion matrix after two-pass classification

From the above discussion, it is evident that the recognition performance improves significantly from 86.2% in the first pass to 90.4% after second pass of classification on 500 training samples of handwritten Bangla digits. Pattern classes 2, 4, 7 and 8 are not found suitable to be included in any of the existing groups or in a new group. Thus the classification decision obtained after coarse classification for these classes are assumed to be final.

The present work is only an initial finding report of a large experimentation. The work may be extended to incorporate n-fold cross validation of results on training and test data sets of larger sizes. Further, improved fitness functions may be designed to improve the recognition performances.

## 5.0 ACKNOWLEDGEMENTS

Authors are thankful to the CMATER and the SRUVM project, C.S.E. Department, Jadavpur University, for providing necessary infrastructure facilities during the progress of the work. One of the authors, Mr. S. Basu, is thankful to the authorities of MCKV Institute of Engineering for kindly permitting him to carry on the research work.